\documentclass{article}
\usepackage{spconf,amsmath,graphicx}
\usepackage{subfig}
\usepackage{amssymb}
\usepackage{bm}
\usepackage{booktabs}
\usepackage{amsthm}

\title{Iterative Block Tensor Singular Value Thresholding for Extraction of Low Rank Component of Image Data}
\name{Longxi Chen \quad Yipeng Liu \quad Ce Zhu}
\address{School of Electronic Engineering / Center for Robotics / Center for Information in Medicine\\
	University of Electronic Science and Technology of China (UESTC), Chengdu, 611731, China\\
		emails: chenlongxi@std.uestc.edu.cn, $ \{ $yipengliu, eczhu$ \} $@uestc.edu.cn}
\begin{document}
\ninept
\maketitle
\begin{abstract}

Tensor principal component analysis (TPCA) is a multi-linear extension of principal component analysis which converts a set of correlated measurements into several principal components. In this paper, we propose a new robust TPCA method to extract the principal components of the multi-way data based on tensor singular value decomposition. The tensor is split into a number of blocks of the same size. The low rank component of each block tensor is extracted using iterative tensor singular value thresholding method. The principal components of the multi-way data are the concatenation of all the low rank components of all the block tensors. We give the block tensor incoherence conditions to guarantee the successful decomposition. This factorization has similar optimality properties to that of low rank matrix derived from singular value decomposition. Experimentally, we demonstrate its effectiveness in two applications, including motion separation for surveillance videos and illumination normalization for face images.\\
\end{abstract}
\section{Introduction}
\label{sec:intro}

The high-dimensional data, also referred to as tensors, arise naturally in a number of scenarios, including image and video processing, and data mining \cite{kolda2008scalable}. However, most of the current processing techniques are developed for two-dimensional data \cite{friedman2001elements}. The principal component analysis (PCA)  is one of the most widely used one in two-dimensional data analysis \cite{Jolliffe2002Principal}.

The robust PCA (RPCA), as an extension of PCA, is an effective method in matrix decomposition problems \cite{Cand2011Robust}. Suppose we have a matrix $ \bm{X}\in \mathbb{R}^{n_{1}\times n_{2}}$ , which can be decomposed as $ \bm{X=L_{0}+S_{0}}$, where $ \bm{L_{0}}$  is the low rank component of the matrix and  $ \bm{S_{0}} $ is the sparse component. The RPCA method has been applied to image alignment \cite{Peng2010RASL}, surveillance video processing \cite{Li2004Statistical}, illumination normalization for face images \cite{Georghiades2001From}. In most applications, the RPCA method should flatten or vectorize the tensor data so as to solve the problem in the matrix. It doesn't use the structural feature of the data effectively since the information loss involves in the operation of matricization.
\begin{figure}[tbp]
	\centering
	\includegraphics[width=220pt, 			      	   keepaspectratio]{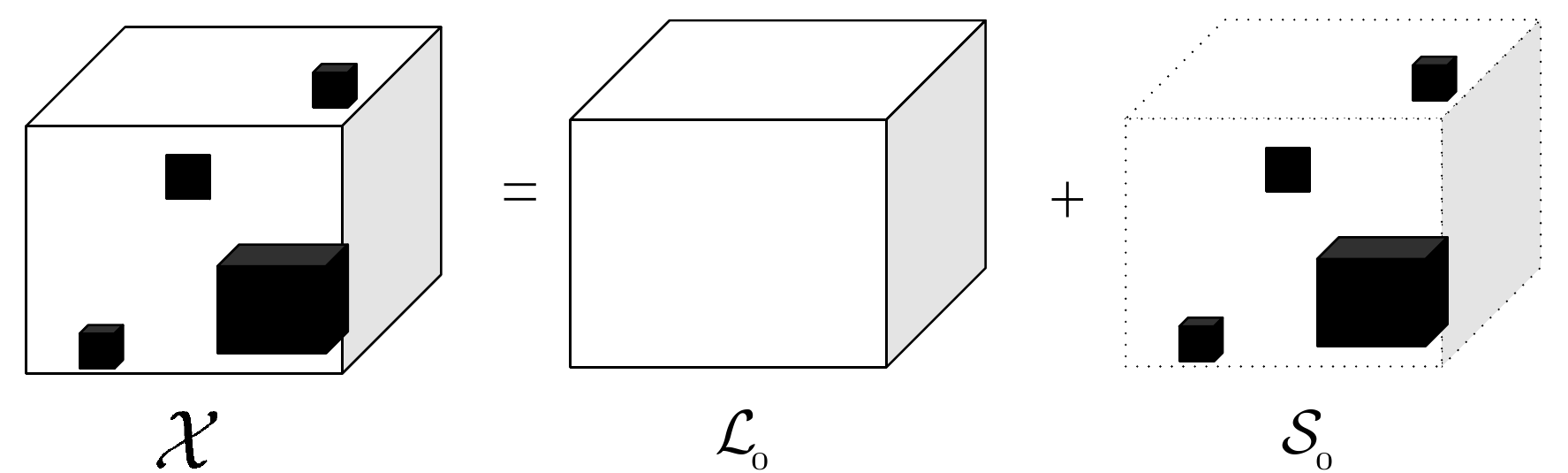}
	\caption{Illustration of TRPCA.}
\end{figure}

Tensor robust principal component analysis (TRPCA) has been studied in \cite{TRPCAcvpr, Zhang2014Novel} based on the tensor singular value decomposition (t-SVD). The advantage of t-SVD over the existing methods such as canonical polyadic decomposition (CPD) \cite{Cichocki2014Tensor} and Tucker decomposition \cite{delathauwer2000a} is that the resulting analysis is very close to that of matrix analysis \cite{Zhang2015Exact}. Similarly, suppose we are given a tensor $ \mathcal{X}\in \mathbb{R}^{n_{1}\times n_{2} \times n_{3}} $  and it can be decomposed into low rank component and sparse component. We can write it as
\begin{equation}
\mathcal{X}=\mathcal{L}_0+\mathcal{S}_0,
\end{equation}
where $ \mathcal{L}_0 $  denotes the low rank component, and $ \mathcal{S}_0 $ is the sparse component of the tensor. Fig. 1 is the illustration for TRPCA. In \cite{TRPCAcvpr} the problem (1) is transformed to the convex optimization model:
\begin{equation}
\min_{\mathcal{L}_0,\mathcal{S}_0} \lVert \mathcal{L}_0 \rVert_{*}+\lambda\lVert \mathcal{S}_0\rVert_{1},\quad\mbox{s.~t. }  \mathcal{X}=\mathcal{L}_0+\mathcal{S}_0,
\end{equation}
where $\lVert \mathcal{L}_0 \rVert_{*} $ is the tensor nuclear norm (see section 2 for the definition), $\lVert \mathcal{S}_0\rVert_{1} $ denotes the $\ell_1$-norm. In the paper \cite{Zhang2014Novel} the problem (1) is transformed to another convex optimization model as:
\begin{equation}
\min_{\mathcal{L}_0,\mathcal{S}_0} \lVert \mathcal{L}_0 \rVert_{*}+\lambda\lVert \mathcal{S}_0\rVert_{1,1,2},\quad\mbox{s.~t. }  \mathcal{X}=\mathcal{L}_0+\mathcal{S}_0,
\end{equation}
where $ \lVert \mathcal{S}_0\rVert_{1,1,2} $ is defined as $ \Sigma_{i,j} \lVert \mathcal{S}_0(i,j,:) \rVert_{\mathrm{F}} $.
The two methods solve the tensor decomposition problem based on the t-SVD.

The low rank and sparse matrix decomposition has been improved by the \cite{Ong2015Beyond}. The main idea is incorporating multi-scale structures with low rank methods. The additional multi-scale structures can obtain a more accurate representation than conventional low rank methods. Inspired by this work, we notice that the sparse component in matrix is block-distributed in some applications, e.g. shadow and motion in videos. For these images we find it is more effective to extract the low rank components in a another smaller scale of image data. Here we try to extract low rank components in block tensor data that is stacked by small scale of image data. And when we decompose the tensor data into many small blocks, it is easy to extract the principal component in some blocks that have few sparse components. We model our tensor data as the concatenation of block tensors instead of solving the RPCA problem as a whole big tensor. Fig. 2 is the illustration of concatenation of block tensors.

Based on the above motivation, we decompose the whole tensor into concatenation of blocks of the same size, then we extract low rank component of each block by minimizing the tubal rank of each block tensor. Fig. 3 is the illustration of our method. And we get low rank component of the whole tensor by concatenating all the low rank components of the block tensors. The proposed method can be used to some conventional image processing problems, including motion separation for surveillance videos (Section 4.1) and illumination normalization for face images (Section 4.2). The results of numerical experiments demonstrate that our method outperforms the existing methods in term of accuracy.

 \begin{figure}[tbp]
 	\centering
 	\includegraphics[width=220pt, height =80pt]{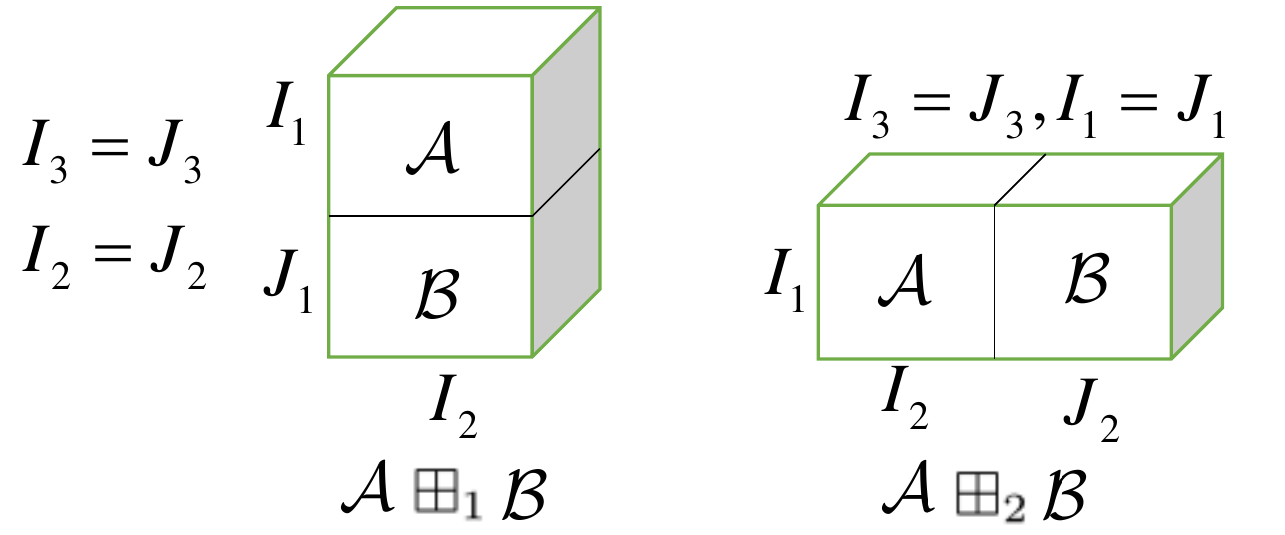}
 	\caption{Illustration of the concatenation of block tensors.}
 \end{figure}

 \begin{figure}[tbp]
 	\centering
 	\includegraphics[width=240pt, 			      	   keepaspectratio]{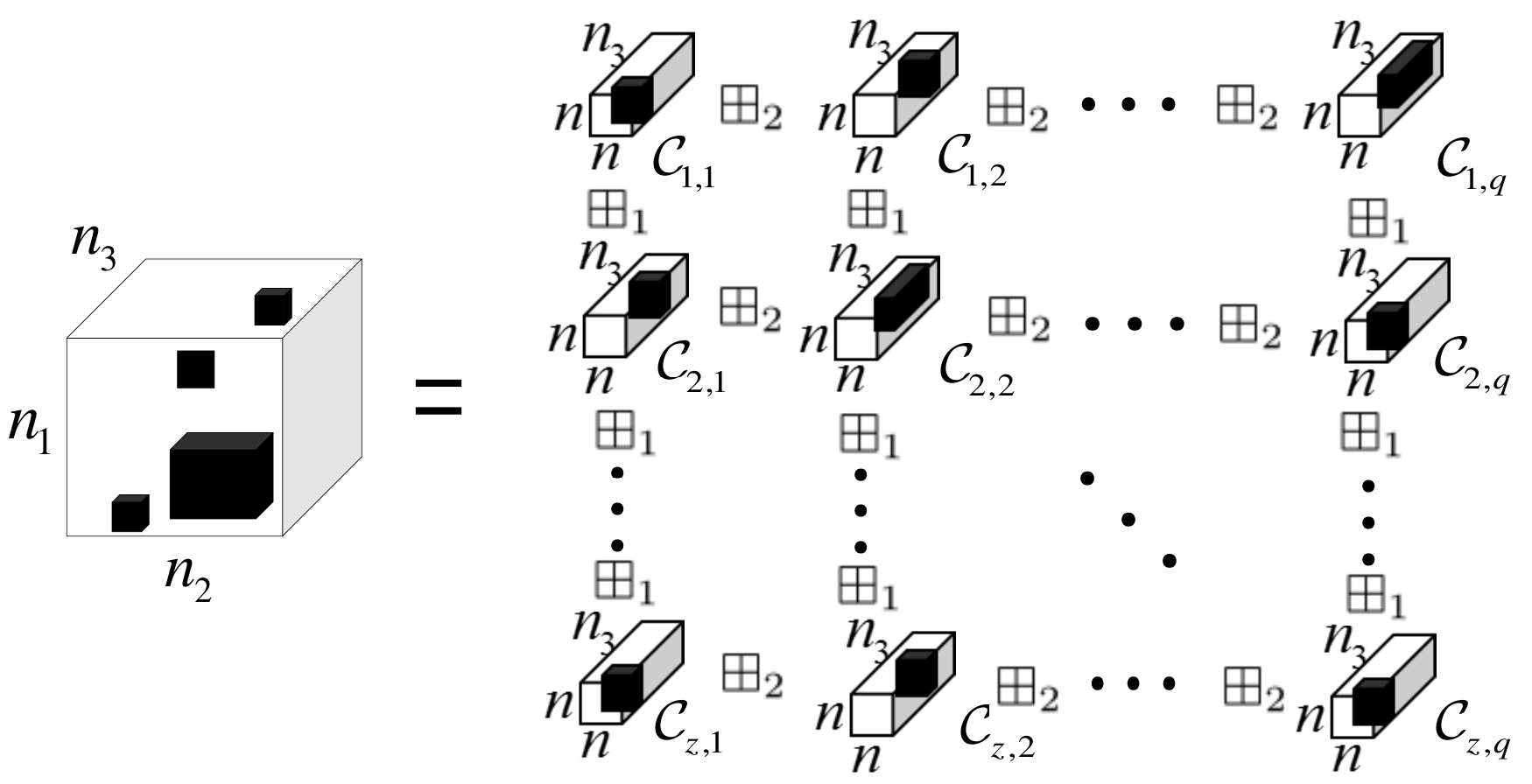}
 	\caption{Illustration of the block tensor decomposition model.}
 \end{figure}

\section{Notations and Preliminaries}
\label{sec:Tensor SVD}
In this section, we describe the notations and definitions used throughout the paper briefly \cite{kilmer2011factorization, Braman2010Third, Kilmer2013Third, Kolda2005Tensor}.

A third-order tensor is represented as $\mathcal{A}$,
and its (\emph{i}, \emph{j}, \emph{k})-th entry is represented as
$ \mathcal{A}_{i,j,k} $. $\mathcal{A}(i,j,:) $ denotes the (\emph{i}, \emph{j})-th tubal scalar. $\mathcal{A}(i,:,:)$, $\mathcal{A}(:,j,:)$ and $\mathcal{A}(:,:,k)$ are  the \emph{i}-th horizontal, \emph{j}-th lateral and \emph{k}-th frontal slices, respectively.$\lVert \mathcal{A} \rVert_\mathrm{F} =\sqrt{\sum_{i,j,k} |a_{ijk}|^2 }  $  and $\lVert \mathcal{A} \rVert_\mathrm{\infty} = \text{max}_{i,j,k}|a_{ijk}|$ tensor kinds of tensor norms. 

We can view a three-dimensional tensor of size  $\emph{n}_1\times\emph{n}_2\times\emph{n}_3$ as an $\emph{n}_1\times\emph{n}_2$ matrix of tubes. $\hat{ \mathcal{A} }$ is a tensor which is obtained by taking the fast Fourier transform  (FFT)  along the third mode of $\mathcal{A}$. For a compact notation we will use
$ \hat{ \mathcal{A} }$ = fft $(\mathcal{A},[],3)$ to denote the FFT along the third dimension. In the same way, we can also compute $\mathcal{A}$ from $\hat{ \mathcal{A} }$ using the inverse FFT (IFFT).\smallskip

\textbf{Definition 2.1} (\emph{\textbf{t-product}}) \emph{\cite{Zhang2015Exact}} \emph{The t-product}  $\mathcal{E}=\mathcal{A} * \mathcal{B}$ \emph{of} $\mathcal{A}\in \mathbb{R}^{\emph{n}_1\times\emph{n}_2\times\emph{n}_3}$  \emph{and} $\mathcal{B}\in \mathbb{R}^{\emph{n}_2\times\emph{n}_4\times\emph{n}_3}$  \emph{is an } $\emph{n}_1\times\emph{n}_4\times\emph{n}_3$  \emph{tensor. The}  $(i,j)$-th  \emph{tube of} $\mathcal{E}$ \emph{is given by}
\begin{equation}
\mathcal{E}(i, j, :)=\sum\limits_{k=1}^{\emph{n}_2}\mathcal{A}(i,k,:) \bullet \mathcal{B}(k,j,:),
\end{equation}
\emph{where $\bullet$  denotes the circular convolution between two tubes of same size}.\smallskip

\textbf{Definition 2.2} (\emph{\textbf{conjugate transpose}}) \emph{\cite{kilmer2011factorization}} \emph{The conjugate transpose of a tensor}  $\mathcal{A}$ \emph{of size} $\emph{n}_1\times\emph{n}_2\times\emph{n}_3$  \emph{ is the} $\emph{n}_2\times\emph{n}_1\times\emph{n}_3$ \emph{ tensor}  $\mathcal{A}^\mathrm{T} $ \emph{ obtained by conjugate transposing each of the frontal slice and then reversing the order of transposed frontal slices from 2 to} $\emph{n}_3$.\smallskip

\textbf{Definition 2.3} (\emph{\textbf{identity tensor}}) \emph{\cite{kilmer2011factorization}} \emph{The identity tensor} $\mathcal{I}\in\mathbb{R}^{\emph{n}\times\emph{n}\times\emph{n}_3}$ \emph{is a tensor whose first frontal slice is the} $\emph{n}\times\emph{n}$  \emph{identity matrix and all other frontal slices are zero}.\smallskip

\textbf{Definition 2.4} (\emph{\textbf{orthogonal tensor}}) \emph{\cite{kilmer2011factorization}} \emph{A tensor $ \mathcal{Q} $ is orthogonal if it satisfies}
\begin{equation}
	\mathcal{Q^\mathrm{T}} * \mathcal{Q}=\mathcal{Q}* \mathcal{Q^\mathrm{T}}=\mathcal{I}.
\end{equation}
\smallskip

\textbf{Definition 2.5} (\emph{\textbf{f-diagonal tensor}}) \emph{\cite{kilmer2011factorization}} \emph{A tensor is called f-diagonal if each of its frontal slices is a diagonal matrix.}\smallskip

\textbf{Definition 2.6} (\emph{\textbf{t-SVD}}) \emph{\cite{kilmer2011factorization}} \emph{For} $\mathcal{A}\in \mathbb{R}^{\emph{n}_1\times\emph{n}_2\times\emph{n}_3}$ , \emph{the t-SVD of}  $\mathcal{A}$ \emph{is given by}
\begin{equation}
\mathcal{A}=\mathcal{U}* \mathcal{S}*\mathcal{V^\mathrm{T}}
\end{equation}
\emph{where $\mathcal{U}$  and $\mathcal{V}$  are orthogonal tensors of size} $\emph{n}_1\times\emph{n}_1\times\emph{n}_3$ \emph{and} $\emph{n}_2\times\emph{n}_2\times\emph{n}_3$  \emph{respectively, and} $\mathcal{S}$  \emph{is a f-diagonal  tensor of size} $\emph{n}_1\times\emph{n}_2\times\emph{n}_3$.\smallskip

We can obtain this decomposition by computing matrix singular value decomposition (SVD) in the Fourier domain, as it shows in Algorithm 1. Fig. 4 illustrates the decomposition for the three-dimensional case.\smallskip

\begin{table}[htbp]
	\centering
	\begin{tabular}{lll}
		\toprule
		$\bf Algorithm 1: $ t-SVD for 3-way data \\
		\midrule
		\textbf{Input:} $ \mathcal A\in \mathbb{R}^{n_{1}\times n_{2}\times n_{3}} $ \\
		
		$\mathcal{D} \gets $fft($\mathcal{A}$,[],$3$),\\
		
		$\bf  for$ $ i=1 $ to $ n_{3}$,  \textbf{do}  \\
		\quad $ [$ \textbf {U , S ,  V }$ ] $ = svd$(\mathcal{D}(:,:,i))$, \\

		\quad$\hat{\mathcal{U}}(:,:,i)$=$ \bf U $,~ $\hat{\mathcal{S}}(:,:,i)$)=$ \bf S $,~ $\hat{\mathcal{V}}(:,:,i)$ =$\bf V $,\\
		\bf end for \\
		
		$\mathcal{U}\gets$ {ifft($\hat{\mathcal{U}}$,[],3)},~ $\mathcal{S}\gets$ ifft($\hat{\mathcal{S}}$,[],3),~ $\mathcal{V}\gets$ ifft($\hat{\mathcal{V}}$,[],3),\\
		
		\textbf{Output:} $\mathcal{U}, \mathcal{S}, \mathcal{V} $\\
		\bottomrule	
	\end{tabular}
\end{table}

 \begin{figure}[htbp]
 	\centering
 	\includegraphics[width=240pt, 			      	   keepaspectratio]{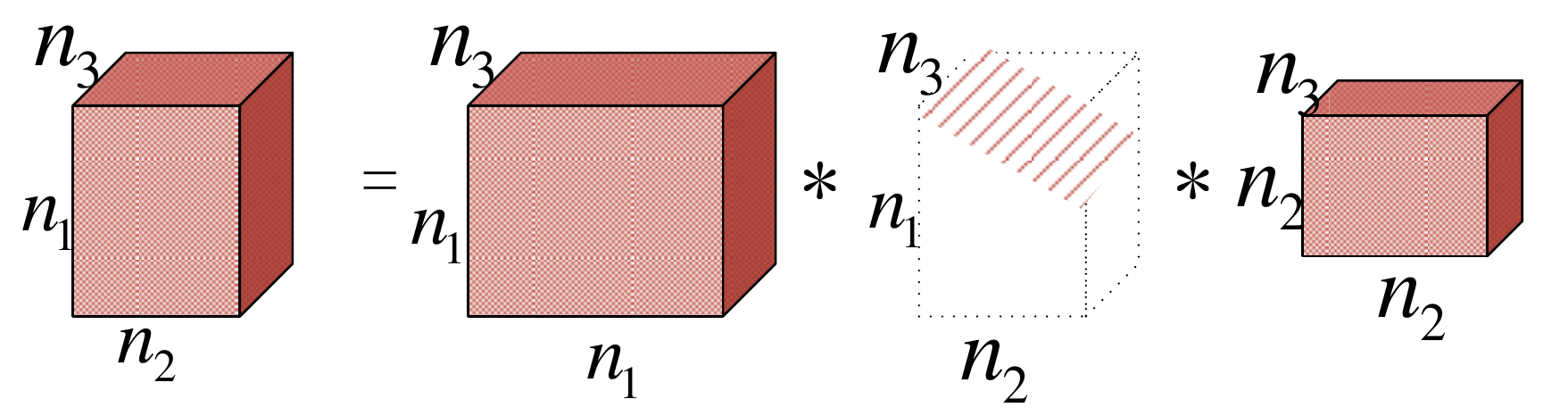}
 	\caption{Illustration of the t-SVD of an $n_1 \times n_2 \times n_3$ tensor.}
 \end{figure}


\textbf{Definition 2.7} (\emph{\textbf{tensor multi-rank and tubal rank}}) \emph{\cite{Zhang2014Novel}} \emph{The tensor multi-rank of} $\mathcal{A}\in \mathbb{R}^{\emph{n}_1\times\emph{n}_2\times\emph{n}_3}$ \emph{is a vector} $ \mathbf{r} \in\mathbb{R}^{\emph{n}_3}$ \emph{with its i-th entry as the rank of the i-th frontal slice of} $\hat{ \mathcal{A} }$, \emph{i. e.} $\emph{r}_i =\mathrm{rank}(\hat{ \mathcal{A}}(:,:,i))$. \emph{The tensor tubal rank, denoted by} {$\mathrm{rank_t}(\mathcal{A})$}, \emph{is defined as the number of nonzero singular tubes of }$\mathcal{S}$, \emph{where} $\mathcal{S}$ \emph{is from } $\mathcal{A}=\mathcal{U}* \mathcal{S}*\mathcal{V^\mathrm{T}}$, \emph{i. e.}
\begin{equation}
	\mathrm{rank_t}(\mathcal{A})=\#\left\{i:\mathcal{S}(\emph{i,i,:})\neq 0\right\}= \max\limits_{i}\emph{r}_i
\end{equation}

\textbf{Definition 2.8} (\emph{\textbf{tensor nuclear norm: TNN}}) \emph{\cite{Zhang2014Novel}} \emph{The tensor nuclear norm of}$\mathcal{A}\in \mathbb{R}^{\emph{n}_1\times\emph{n}_2\times\emph{n}_3}$ \emph{denoted by} $\|\mathcal{A}\|_*$ \emph{ is defined as the sum of the singular values of all the frontal slices of }$\hat{ \mathcal{A} }$. \emph{The TNN of }$\mathcal{A}$ \emph{ is equal to the TNN of }\emph{blkdiag($\hat{ \mathcal{A} }$)}. \emph{Here \text{blkdiag}($\hat{ \mathcal{A} }$)} \emph{is a block diagonal matrix defined as follows:}
\begin{equation}
	\mbox{blkdiag}(\hat{ \mathcal{A} })=
	\left[
	\begin{array}{cccc}
		\hat{ \mathcal{A} }^{(1)} & & &\\
		& \hat{ \mathcal{A} }^{(2)}& &\\
		& & \ddots& \\
		& & &\hat{ \mathcal{A} }^{(\emph{n}_3)}
	\end{array}
	\right],
\end{equation}
\emph{where} $\hat{ \mathcal{A} }^{(i)}$ \emph{is the i}-th
\emph{frontal slice of }$\hat{ \mathcal{A} }, i=1,2,...,n_3$.\smallskip

\textbf{Definition 2.9} (\emph{\textbf{standard tensor basis}}) \emph{\cite{Zhang2015Exact}} \emph{The column basis, denoted as }$\mathring {\mathfrak{e}}_i$, \emph {is a tensor of size }$\emph{n}\times{1}\times\emph{n}_3$ \emph{with its} (\emph{i},1,1)-th  \emph{entry equaling to 1} \emph{and the rest equaling to 0}. \emph{Naturally its transpose $\mathring {\mathfrak{e}}_i^\mathrm{T}$  is called row basis.}

\section{Iterative Block Tensor Singular Value Thresholding}
\label{Our method about Tensor PCA}

We decompose the whole tensor which satisfies the incoherence conditions into many small blocks of the same size. And the third dimension of the block size should be the same as the third dimension of the tensor. That is to say, given an input tensor $\mathcal{X}\in\mathbb{R}^{\emph{n}_1\times\emph{n}_2\times\emph{n}_3}$ and its corresponding block size, we propose a multi-block tensor modeling that models the tensor data $\mathcal{X}$ as the concatenation of block tensors. And each block tensor can be decomposed into two components, i.e.
$ \mathcal{X}_{p}=\mathcal{L}_{p}+\mathcal{S}_{p},~p = 1, \cdots , P $, where $\mathcal{L}_{p}$ and $\mathcal{S}_{p}$ denote the low rank component and sparse component of block tensor $ \mathcal{X}_p$ respectively.

As observed in RPCA, the low rank and sparse decomposition is impossible in some cases \cite{Cand2011Robust}. Similarly, we are not able to identify the low rank component and sparse component if the tensor is of both low rank and sparsity. Similar to the tensor incoherence conditions \cite{TRPCAcvpr}, we assume the block tensor data $ \mathcal{L}_p $ in each block satisfies some block tensor incoherence conditions to guarantee successful low rank component extraction.

\textbf{Definition 3.1} (\emph{\textbf{block tensor incoherence conditions}})
\emph{For} $\mathcal{L}_p\in\mathbb{R}^{\emph{n}\times\emph{n}\times\emph{n}_3}$, \emph{ assume that}  $ \mathrm{rank_t}(\mathcal{L}_p)=r $ \emph {and it has the t-SVD }$\mathcal{L}_p=\mathcal{U}_p* \mathcal{S}_p*\mathcal{V}_p^\mathrm{T}$, \emph{where} $\mathcal{U}_p\in \mathbb{R}^{\emph{n}\times\emph{r}\times\emph{n}_3}$ \emph{and}
$\mathcal{V}_p\in \mathbb{R}^{\emph{n}\times\emph{r}\times\emph{n}_3}$ \emph{satisfy} $\mathcal{U}_p^\mathrm{T} * \mathcal{U}_p=\mathcal{I}$ \emph{and} $\mathcal{V}_p^\mathrm{T} * \mathcal{V}_p=\mathcal{I}$, \emph{and} $\mathcal{S}_p\in\mathbb{R}^{\emph{r}\times\emph{r}\times\emph{n}_3}$ \emph{ is an f-diagonal tensor}. \emph{Then} $\mathcal{L}_p$ \emph{satisfies the tensor incoherence conditions with parameter }$ \mu $ \emph{if}
\begin{equation}
\max_{i=1,...,n} \lVert \mathcal{U}_p^\mathrm{T} * \mathring{\mathfrak{e}}_i   \rVert_\mathrm{F} \leqslant \sqrt{\frac{\mu{r}}{nn_3}}
\end{equation}
\begin{equation}
\max_{j=1,...,n} \lVert \mathcal{V}_p^\mathrm{T} * \mathring{\mathfrak{e}}_j   \rVert_\mathrm{F} \leqslant \sqrt{\frac{\mu{r}}{nn_3}}
\end{equation}
and
\begin{equation}
	\lVert \mathcal{U}_p * \mathcal{V}_p^\mathrm{T}\rVert_\mathrm{\infty} \leqslant \sqrt{\frac{\mu{r}}{n^2n_3^2}}
\end{equation}
The incoherence condition guarantees that for small values of $\mu$, the singular vectors are not sparse. Then the tensor $\mathcal{L}_p\in\mathbb{R}^{\emph{n}\times\emph{n}\times\emph{n}_3}$  can be decomposed into low rank component and sparse component.

For extracting the low rank component from every block, we process the tensor nuclear norm of $ \mathcal{L}_{p}$, i. e. $ \lVert \mathcal{L}_{p} \rVert_\mathrm{TNN}= \lVert \mbox{blkdiag}(\hat{\mathcal{L}_{p}}) \rVert_*$. Here we can use singular value thresholding operator in the Fourier domain to extract the low rank component of the block tensor \cite{Cai2008A, Watson1992Characterization}. The proposed method is called iterative block tensor singular value thresholding (IBTSVT).
The thresholding operator used here is the soft one $D_\tau$ as follows:
\begin{equation}
D_\tau(\mathcal{L}_{p}) = \mathrm{sign}(\mbox{blkdiag}(\hat{\mathcal{L}}_{p}))(|\mbox{blkdiag}(\hat{\mathcal{L}_{p}})| - \tau )_+,
\end{equation}
where ``$()_+$" keeps the positive part.

After we extract the low rank component $ \mathcal{L} = \mathcal{L}_1\boxplus \mathcal{L}_2\boxplus \cdots \boxplus \mathcal{L}_P$, where $ \boxplus $ denotes concatenation operation, we can get the sparse component of the tensor by computing the $\mathcal{S} =\mathcal{X}- \mathcal{L} $. See Algorithm 2 in details.

\begin{table}[htbp]
	\centering
	\begin{tabular}{lll}
		\toprule
		$\bf Algorithm~2 $: IBTSVT \\
		\midrule
		$ \bf Input$: \mbox{tensor data} $ \mathcal X\in \mathbb{R}^{n_{1}\times n_{2}\times n_{3}}$\\
		\textbf {Initialize}: given $\mu$, $\eta$, $\epsilon$, $ \tau$, and
         \\ block tensors $ \mathcal{X}_p$ of  size $n\times n\times n_3$, $ p =1, \cdots, P $\\
		\textbf{while} not converged \textbf{do}\\
		1. Update $\eta := \eta \times \mu $,\\
		2. Update $\tau := \tau/\eta$,\\
		3. Compute $\mathcal{X}_p := D_{\tau}({\mathcal{X}_p})$, $ p =1, \cdots, P $.\\
				
		\bf end while:
        \\ $\lVert \mathcal{X}_{k+1}-\mathcal{X}_k \rVert_\text{F}/\lVert \mathcal{X}_k \rVert_\text{F} \leq \epsilon$ at ($k+1$)-th step.\\
		\bf Output: $ \mathcal{L} = \mathcal{X}_1\boxplus \mathcal{X}_2\boxplus \cdots \boxplus \mathcal{X}_P$ \\
		\bottomrule	
	\end{tabular}
\end{table}

In our method, the block size can't be too large. The large size of the block will make the sparse part contain some low rank component. And if the size of the block is too small, the computational time will be long. Because the number of t-SVDs is large. Generally, we can choose our block size $2 \times 2 \times n_3$.

In our algorithm, we choose $\mu=1.8$, $\eta=1$, $\epsilon=10^{-2}$. But the thresholding parameter $ \tau $ is difficult to determine. Here we can get a value by experience. As discussed in \cite{TRPCAcvpr}, the thresholding parameter could be $ \tau = 1/\sqrt{nn_3}$  for every block. This value is for denoising problem in images or videos, where the noise is uniformly distributed. But for different applications, it should be different from  $1/\sqrt{nn_3}$. Because in these applications, the sparse component in data is not uniformly distributed, such as shadow in face images and motion in surveillance videos.

\section{Experimental Results}
\label{sec:experiment}
In this section, we conduct numerical results to show the performance of the method. We apply IBTSVT method on two different real datasets that are conventionally used in low rank model: motion separation for surveillance videos (Section 4.1) and illumination normalization for face images (Section 4.2).

\subsection{Motion Separation for Surveillance Videos}
\label{sec:Motion Separation}
In surveillance video, the background only changes its brightness over the time, and it can be represented as the low rank component. And the foreground objects are the sparse component in videos. It is often desired to extract the foreground objects from the video. We use the proposed IBTSVT method to separate the foreground component from the background one.

We use the surveillance video data used in \cite{Li2004Statistical}. Each frame is of size $144 \times 176$  and we use 20 frames. The constructed tensor is  $\mathcal{X}\in\mathbb{R}^{144\times 176\times 20}$ and the selected block size is $2 \times 2 \times 20$. The thresholding parameter is $ \tau = 20/\sqrt{nn_3}$.

Fig. 5 shows one of the results. We can find that IBTSVT method correctly recovers the background, while the sparse component correctly identifies the moving pedestrians. It shows the proposed method can realize motion separation for surveillance videos.

 \begin{figure}[h]
 	\centering
 	\includegraphics[width=220pt, 			      	   keepaspectratio]{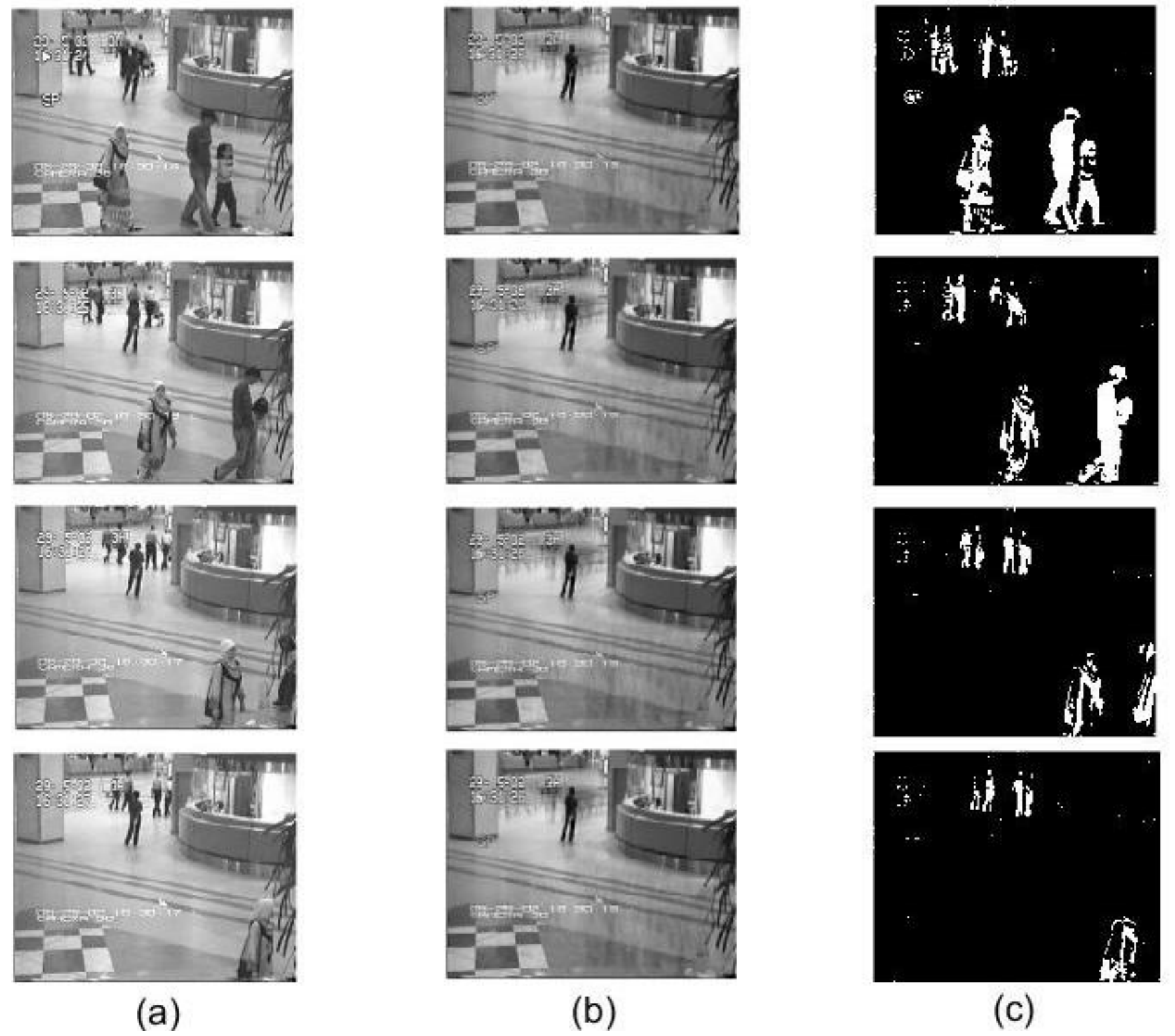}
 	\caption{IBTSVT on a surveillance video. (a) original video; (b) low rank component that is video background; (c) sparse component that represents the foreground objects of video.}
 \end{figure}

\subsection{Illumination normalization for face images}
\label{sec:illumination}
The face recognition algorithms are sensitive to shadow or occlusions on faces \cite{Georghiades2001From}, and it's important to remove illumination variations and shadow on the face images. The low rank model is often used for face images \cite{Basri2003Lambertian}.
 
In our experiments, we use the Yale B face database \cite{Georghiades2001From}. Each face image is of size $192 \times 168$  with 64 different lighting conditions. We construct the tensor data $\mathcal{X}\in\mathbb{R}^{192\times 168\times 64}$  and choose the block size $2 \times 2 \times 64$  . We set the thresholding parameter  $ \tau = 20/\sqrt{nn_3}$.

We compare the proposed method with multi-scale low rank matrix decomposition method \cite{Ong2015Beyond} and low rank + sparse method \cite{Cand2011Robust}. Fig. 6 shows one of the comparison results. The IBTSVT method can result in almost shadow-free faces. In contrast, the other two methods can only recover the faces with some shadow.

In order to further illustrate the effect of shadow elimination in the recovered face images, we carry on face detection with the recovered data from different methods. In our experiments, we employ the face detection algorithm Viola-Jones algorithm  \cite{Viola2001Rapid} to detect the faces and the eyes. The Viola-Jones algorithm is a classical algorithm which can be used to detect people's faces, noses, eyes, mouths, and upper bodies. In the first experiment we put all face images into one image of JPG format. Then we use the algorithm to detect faces in the newly formed image. In the second experiment, we use the algorithm to detect the eyes of every face image. The second and third columns of Table 1 show the detection accuracy ratios of Viola-Jones algorithm with different recovered face images. We test how long the three methods process the 64 face images as can be seen in the fourth column. The IBTSVT can improve the efficiency by parallel processing of the block tensors. From the result of Table 1, we can find our method gives the best detection performance, because removing shadow of face images is helpful for face detection.

 \begin{figure}[h]
 	\centering
 	\includegraphics[width=220pt, 			      	   keepaspectratio]{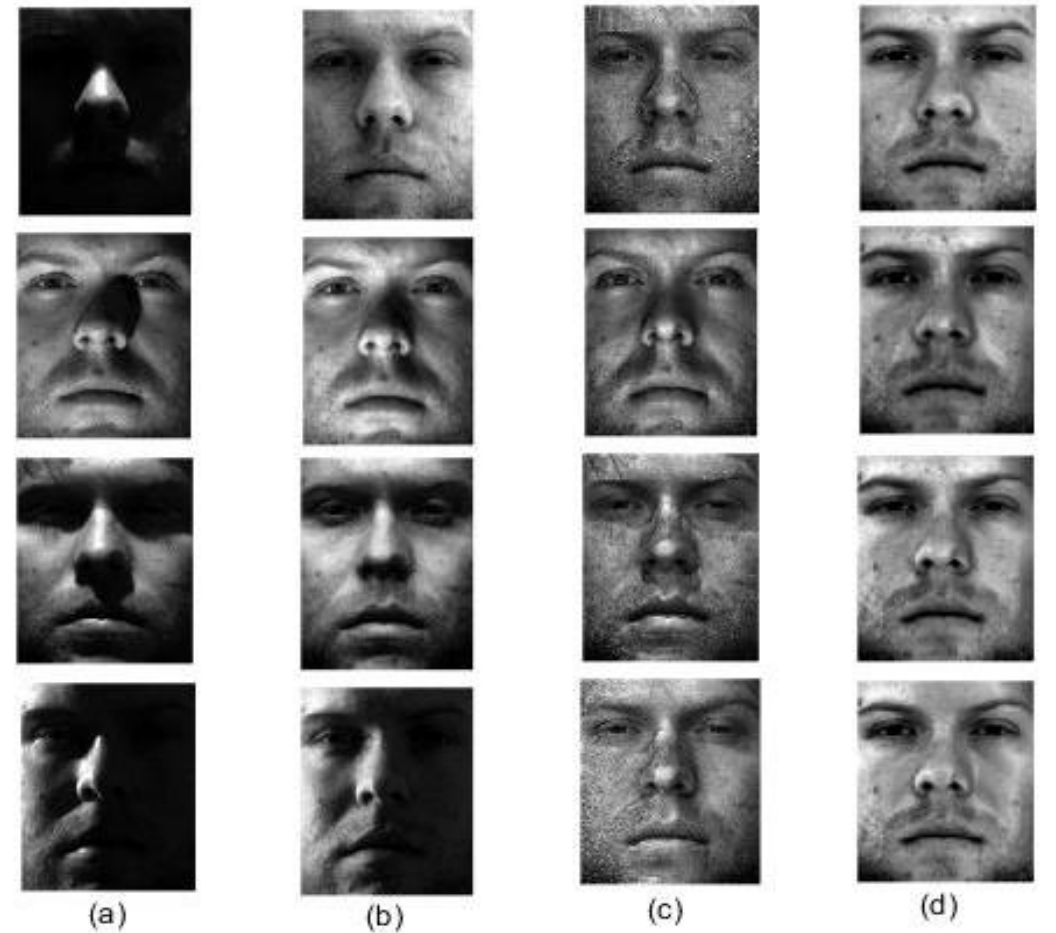}
 	\caption{Three methods for face with uneven illumination: (a) original faces with shadows; (b)  low rank + sparse  method; (c) multi-scale low rank decomposition; (d) IBTSVT.}
 \end{figure}

\begin{table}[h]
	\centering
	\begin{tabular}{|l|c|c|r|}
		\hline
		&face detection & eye detection & Time (s)\\
		\hline
		Original image & 0.297 & 	0.58 & NULL\\
		\hline
		Low rank + sparsity & 0.375 & 0.70 & 10 \\
		\hline
		Multiscale low rank & 0.359 & 1.00 & 4472\\
		\hline
		IBTSVT& \textbf{0.844} & \textbf{1.00} & \textbf{715}\\
		\hline
	\end{tabular}
	\caption{The accuracy ratios of faces and eyes detection by Viola-Jones algorithm and the computational time to process face images.}
\end{table}

\section{Conclusions}
\label{sec:conclusion}
In this paper, we proposed a novel IBTSVT method to extract the low rank component of the tensor using t-SVD. The IBTSVT is a good way to utilize the structural feature of tensor by solving TPCA problem in block tensor form. We have given the tensor incoherence conditions for block tensor data. For applications, we considered motion separation for surveillance videos and illumination normalization for face images, and numerical experiments showed its performance gains compared with the existing methods.

\section{ACKNOWLEDGMENT}
\label{sec:acknowledgment}
This work is supported by National High Technology Research and Development Program of China (863, No. 2015AA015903), National Natural Science Foundation of China (NSFC, No. 61602091, No. 61602091), the Fundamental Research Funds for the Central Universities (No. ZYGX2015KYQD004, No. ZYGX2015KYQD004), and a grant from the Ph.D. Programs Foundation of Ministry of Education of China (No. 20130185110010).

\bibliographystyle{IEEE}
\bibliography{citation}

\begin{thebibliography}{10}

\bibitem{kolda2008scalable}
T.~G. Kolda and J.~Sun,
\newblock ``Scalable tensor decompositions for multi-aspect data mining,''
\newblock in {\em IEEE International Conference on Data Mining}. IEEE, 2008,
  pp. 363--372.

\bibitem{friedman2001elements}
J.~Friedman, T.~Hastie, and R.~Tibshirani,
\newblock {\em The elements of statistical learning}, vol.~1,
\newblock Springer series in statistics Springer, Berlin, 2001.

\bibitem{Jolliffe2002Principal}
I.~Jolliffe,
\newblock {\em Principal Component Analysis},
\newblock New York: Springer, 2002.

\bibitem{Cand2011Robust}
E.~J. Cand{\`e}s, X.~Li, Y.~Ma, and J.~Wright,
\newblock ``Robust principal component analysis,''
\newblock {\em Journal of the ACM}, vol. 58, no. 3, pp. 1--73, 2011.

\bibitem{Peng2010RASL}
Y.~Peng, A.~Ganesh, J.~Wright, W.~Xu, and Y.~Ma,
\newblock ``{RASL}: robust alignment by sparse and low-rank decomposition for
  linearly correlated images,''
\newblock {\em IEEE Transactions on Pattern Analysis and Machine Intelligence},
  vol. 34, no. 11, pp. 2233--2246, 2010.

\bibitem{Li2004Statistical}
L.~Li, W.~Huang, I.~Y.~H. Gu, and Q.~Tian,
\newblock ``Statistical modeling of complex backgrounds for foreground object
  detection,''
\newblock {\em IEEE Transactions on Image Processing}, vol. 13, no. 11, pp.
  1459--72, 2004.

\bibitem{Georghiades2001From}
A.~S. Georghiades, P.~N. Belhumeur, and D.~J. Kriegman,
\newblock ``From few to many: Illumination cone models for face recognition
  under variable lighting and pose,''
\newblock {\em IEEE Transactions on Pattern Analysis and Machine Intelligence},
  vol. 23, no. 6, pp. 643--660, 2001.

\bibitem{TRPCAcvpr}
C.~Lu, Y.~Chen J.~Feng, W.~Liu, Z.~Lin, and S.~Yan,
\newblock ``Tensor robust principal component analysis: Exact recovery of
  corrupted low-rank tensors via convex optimization,''
\newblock in {\em The IEEE Conference on Computer Vision and Pattern
  Recognition}, June 2016.

\bibitem{Zhang2014Novel}
Z.~Zhang, G.~Ely, S.~Aeron, N.~Hao, and M.~Kilmer,
\newblock ``Novel methods for multilinear data completion and de-noising based
  on tensor-svd,''
\newblock {\em Computer Science}, vol. 44, no. 9, pp. 3842--3849, 2014.

\bibitem{Cichocki2014Tensor}
A.~Cichocki, D.~Mandic, L.~De Lathauwer, G.~Zhou, Q.~Zhao, C.~Caiafa, and H.~A.
  Phan,
\newblock ``Tensor decompositions for signal processing applications: From
  two-way to multiway component analysis,''
\newblock {\em IEEE Signal Processing Magazine}, vol. 32, no. 2, pp. 145--163,
  2014.

\bibitem{delathauwer2000a}
L.~De Lathauwer, B.~De Moor, and J.~Vandewalle,
\newblock ``A multilinear singular value decomposition,''
\newblock {\em SIAM Journal on Matrix Analysis and Applications}, vol. 21, no.
  4, pp. 1253--1278, 2000.

\bibitem{Zhang2015Exact}
Z.~Zhang and S.~Aeron,
\newblock ``Exact tensor completion using t-svd,''
\newblock {\em arXiv preprint arXiv:1502.04689}, 2015.

\bibitem{Ong2015Beyond}
F.~Ong and M.~Lustig,
\newblock ``Beyond low rank + sparse: Multiscale low rank matrix
  decomposition,''
\newblock {\em IEEE Journal of Selected Topics in Signal Processing}, vol. 10,
  no. 4, pp. 672--687, 2015.

\bibitem{kilmer2011factorization}
M.~E. Kilmer and C.~D. Martin,
\newblock ``Factorization strategies for third-order tensors,''
\newblock {\em Linear Algebra and its Applications}, vol. 435, no. 3, pp.
  641--658, 2011.

\bibitem{Braman2010Third}
K.~Braman,
\newblock ``Third-order tensors as linear operators on a space of matrices,''
\newblock {\em Linear Algebra and Its Applications}, vol. 433, no. 7, pp.
  1241--1253, 2010.

\bibitem{Kilmer2013Third}
M.~E. Kilmer, E.~Misha, K.~Braman, N.~Hao, and R.~C. Hoover,
\newblock ``Third-order tensors as operators on matrices: a theoretical and
  computational framework with applications in imaging,''
\newblock {\em SIAM Journal on Matrix Analysis and Applications}, vol. 34, no.
  1, pp. 148--172, 2013.

\bibitem{Kolda2005Tensor}
T.~G. Kolda and B.~W. Bader,
\newblock ``Tensor decompositions and applications,''
\newblock {\em SIAM Review}, vol. 66, no. 4, pp. 294--310, 2005.

\bibitem{Cai2008A}
J.~F. Cai, E.~J. Cand{\`e}s, and Z.~Shen,
\newblock ``A singular value thresholding algorithm for matrix completion,''
\newblock {\em SIAM Journal on Optimization}, vol. 20, no. 4, pp. 1956--1982,
  2008.

\bibitem{Watson1992Characterization}
G.~A. Watson,
\newblock ``Characterization of the subdifferential of some matrix norms,''
\newblock {\em Linear Algebra and Its Applications}, vol. 170, no. 6, pp.
  33--45, 1992.

\bibitem{Basri2003Lambertian}
R.~Basri and D.~W. Jacobs,
\newblock ``Lambertian reflectance and linear subspaces,''
\newblock {\em IEEE Transactions on Pattern Analysis and Machine Intelligence},
  vol. 25, no. 2, pp. 218--233, 2003.

\bibitem{Viola2001Rapid}
P.~Viola and M.~Jones,
\newblock ``Rapid object detection using a boosted cascade of simple
  features,''
\newblock in {\em Computer Vision and Pattern Recognition}. IEEE, 2001, vol.~1,
  pp. I--511.

\end{thebibliography}
\label{sec:refs}
\end{document}